\documentclass[letterpaper,10pt,conference,transmag]{IEEEtran}
\IEEEoverridecommandlockouts

\def\IEEEtitletopspace{18pt}

\usepackage{cite}
\usepackage{amsmath,amssymb,amsfonts}
\usepackage{algorithmic}
\usepackage{graphicx}
\usepackage{subcaption}
\usepackage{textcomp}
\usepackage{xcolor}

\usepackage{booktabs}       
\usepackage{multirow}
\usepackage{makecell}

\usepackage[hidelinks,breaklinks=true]{hyperref}

\usepackage{soul}
\usepackage{bm}

\usepackage{array}

\newcommand{\synthetic}{\texttt{synthetic}}
\newcommand{\lightbox}{\texttt{lightbox}}
\newcommand{\sunlamp}{\texttt{sunlamp}}
\newcommand{\prisma}{\texttt{prisma25}}

\newcommand{\mysubsection}[1]{{\bf#1.}~}

\newcommand{\repv}[2]{\makecell{#1 \\ \tiny{(#2)}}}

\newcommand{\link}[1]{{\color{blue}\ttfamily\url{#1}}}

\begin{document}

\title{Online Supervised Training of Spaceborne Vision during Proximity Operations using Adaptive Kalman Filtering \\ 
\thanks{The authors are with the Space Rendezvous Laboratory (SLAB), Department of Aeronautics \& Astronautics, Stanford University, Stanford, CA 94305. Email: \texttt{\{tpark94, damicos\}@stanford.edu}.}
\thanks{For visualization of various datasets and results, please see the project website: \link{https://taehajeffpark.com/ost/}.}
\thanks{This work is supported by the US Space Force SpaceWERX Orbital Prime Small Business Technology Transfer (STTR) contract number FA8649-23-P-0560 awarded to TenOne Aerospace in collaboration with SLAB.}
}

\author{\IEEEauthorblockN{Tae Ha Park ~~~ Simone D'Amico}
}

\IEEEaftertitletext{\vspace{-0.5\baselineskip}}

\maketitle

\begin{abstract}
This work presents an Online Supervised Training (OST) method to enable robust vision-based navigation about a non-cooperative spacecraft. Spaceborne Neural Networks (NN) are susceptible to domain gap as they are primarily trained with synthetic images due to the inaccessibility of space. OST aims to close this gap by training a pose estimation NN online using incoming flight images during Rendezvous and Proximity Operations (RPO). The pseudo-labels are provided by an adaptive unscented Kalman filter where the NN is used in the loop as a measurement module. Specifically, the filter tracks the target's relative orbital and attitude motion, and its accuracy is ensured by robust on-ground training of the NN using only synthetic data. The experiments on real hardware-in-the-loop trajectory images show that OST can improve the NN performance on the target image domain given that OST is performed on images of the target viewed from a diverse set of directions during RPO.
\end{abstract}


\section{Introduction} \label{sec:intro}

In space robotics, one of the most sought-after capabilities is autonomous Guidance, Navigation and Control (GN\&C) with respect to non-cooperative objects such as satellites and debris. This requires estimating the pose (i.e., position and orientation) of the target relative to the servicer, which enables various future missions for on-orbit servicing \cite{reed_2016_aiaa_restorel} and debris removal \cite{forshaw_2016_acta_removedebris}. Performing pose estimation using a single monocular camera is particularly attractive due to its low mass and power requirements suitable for on-board satellite avionics.

Machine Learning (ML) and Neural Networks (NN) have recently emerged as prevailing methods for pose estimation especially for known targets \cite{park_2019_aas, kisantal_2020_taes_spec2019, sharma_2020_taes_spn, chen_2019_iccvw_pose, proenca_2020_icra_urso, kaki_2023_jais_pose, park_2023_spnv2}. This is a scenario suitable for space missions with a priori information on the client's target spacecraft. However, unlike in terrestrial applications such as autonomous driving, spaceborne NNs must address several challenges unique to space environments. First, the computational resource is scarce on-board the satellite with no or minimal GPU support. Second, access to space is severely restricted, even more so beyond Low Earth Orbit (LEO), which implies a lack of annotated flight images of the target spacecraft for training and validation of NNs. Therefore, spaceborne NNs must instead be trained on synthetic imagery rendered with computer graphics which lack visual characteristics typical of the space imagery. It also implies a lack of representative images for \emph{on-ground} performance validation which is crucial for ensuring the success of costly and safety-critical space missions.

\begin{figure}[!t]
\centering
\begin{subfigure}[c]{0.24\textwidth}
    \centering
    \includegraphics[width=\textwidth]{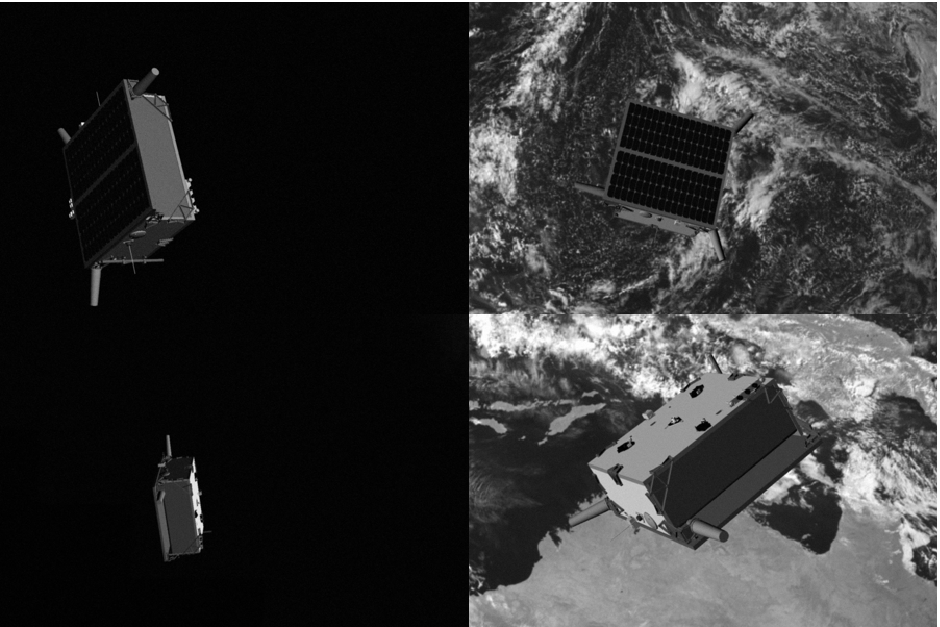}
    \caption{$\synthetic$}
\end{subfigure}
\begin{subfigure}[c]{0.24\textwidth}
    \centering
    \includegraphics[width=\textwidth]{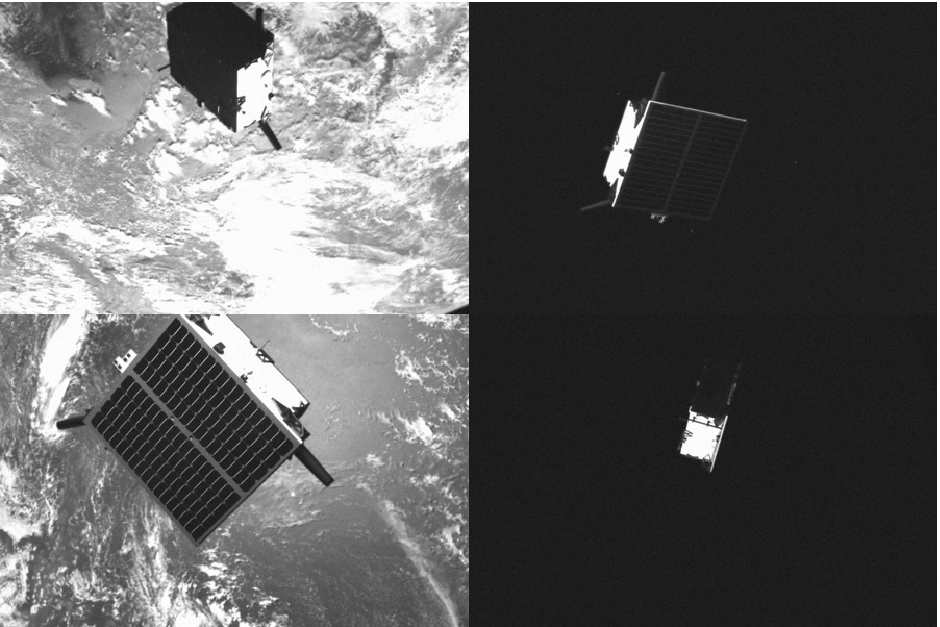}
    \caption{$\lightbox$}
\end{subfigure}
\begin{subfigure}[c]{0.24\textwidth}
    \centering
    \includegraphics[width=\textwidth]{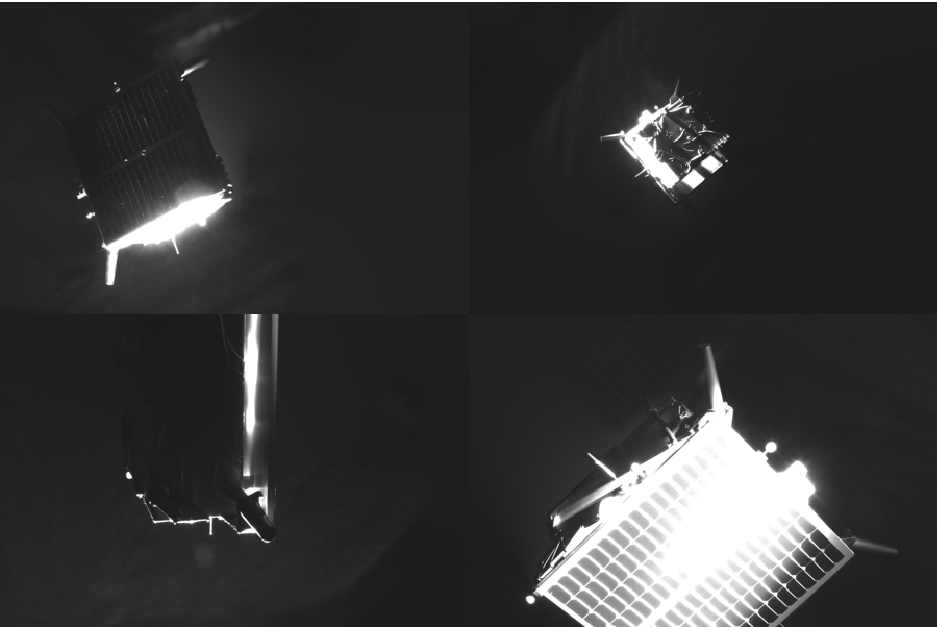}
    \caption{$\sunlamp$}
\end{subfigure}
\begin{subfigure}[c]{0.24\textwidth}
    \centering
    \includegraphics[width=\textwidth]{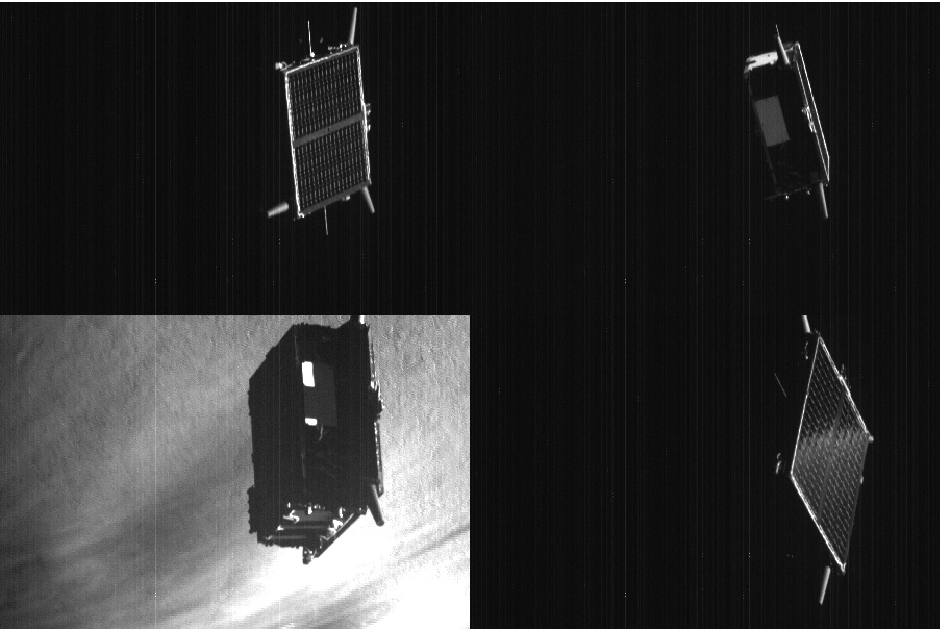}
    \caption{$\prisma$}
\end{subfigure}
\caption{Various image domains of the Tango spacecraft from the SPEED+ dataset \mbox{\cite{park_2022_speedplus}} (a, b, c) and actual space flight (d).}
\label{fig:image domains}
\end{figure}

As flight images do not become available until close-range rendezvous in space, one key strategy is to instead create high-fidelity surrogate images on-ground that can be used to evaluate the robustness of NN models trained on synthetic images across domain gap (also known as sim2real gap) on flight images. These so-called Hardware-In-the-Loop (HIL) images can be created along with high-accuracy pose labels using a satellite mockup model in a robotic testbed capable of emulating high-fidelity space illumination conditions \cite{park_2021_aas_tron}. For instance, Fig.~\ref{fig:image domains} shows examples of HIL image domains of the SPEED+ dataset \cite{park_2022_speedplus}, $\lightbox$ and $\sunlamp$, which comprise images of the mockup model of the Tango spacecraft from the PRISMA mission \cite{prisma_chapter, gill_2007_jgcd_prisma} illuminated with albedo light boxes and a sun lamp, respectively. Captured with an actual camera inside a realistic space-like environment, these HIL images can be used for comprehensive evaluation of NN robustness on otherwise unavailable flight images. However, while HIL images can significantly reduce the visual gap between the synthetic and spaceborne images, there are still remaining gaps largely due to the fact that HIL domains create images of an inexpensive mockup model that fails to capture the real satellite's material and surface properties.

\begin{figure}[!t]
\centering
    \centering
    \includegraphics[width=0.47\textwidth]{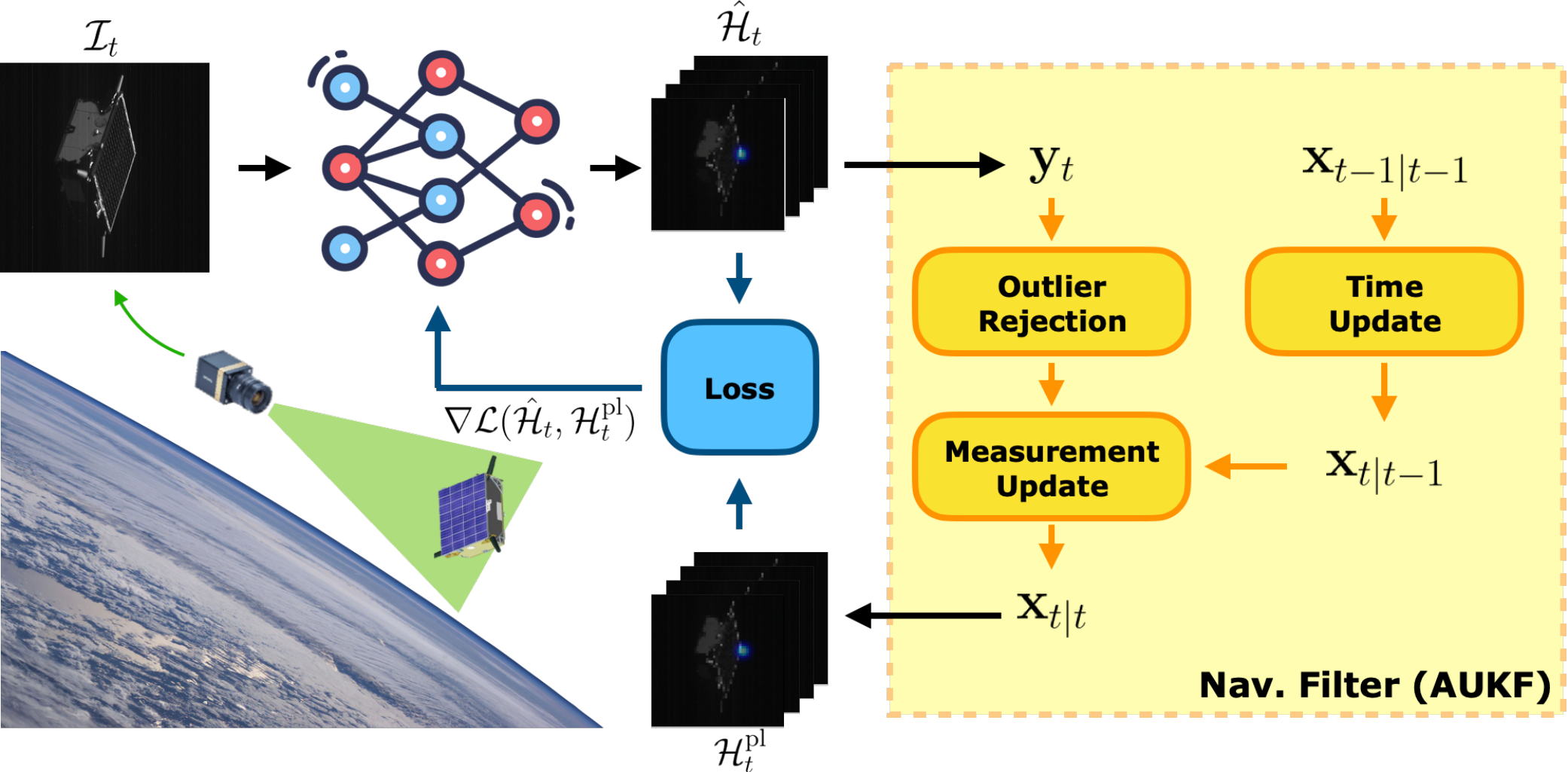}
    \caption{Illustration of OST using an Adaptive Unscented Kalman Filter (AUKF) during space RPO.}
    \label{fig:pipeline}
\end{figure}

In order to completely close the domain gap, this work proposes to perform Online Supervised Training (OST) using incoming flight images in space during Rendezvous and Proximity Operations (RPO). As shown in Fig.~\ref{fig:pipeline}, the pose estimation NN is integrated into the navigation filter as a measurement module. The filter tracks the orbital and rotational motion of the known target relative to the servicer, so the state estimates can be used to create pose Pseudo-Labels (PL) for each flight image to perform OST. In order to generate accurate PLs, the NN must be trained and validated on-ground to be robust on unseen flight images. Therefore, a lightweight NN is trained with heavy data augmentation to bias the network away from features specific to synthetic imagery and render it robust across domain gaps on HIL and flight images. This work shows that with enough diversity in image acquisition pose, OST can improve the steady-state errors of the navigation filter even with sub-optimally trained NNs and additionally improve the NN's overall pose estimation capabilities on the test domain as a whole.

\section{Related Work}

\mysubsection{Monocular Spacecraft Pose Estimation} Traditional approaches to monocular pose estimation of known targets in space \cite{cropp_2002_pose, leinz_2008_isop_orbitalexpress_arcss, damico_2014_ijsse_pose, grompone_2015_thesis_pose, sharma_2018_jsr_pose, capuano_2019_scitech_robustfeatures} relied on feature extraction (e.g., edges \cite{canny_1986_edges}, SIFT \cite{lowe_2004_sift}), solving for the optimal 2D-3D correspondence and recovering pose via Perspective-$n$-Point (P$n$P) \cite{lepetit_2008_epnp, sharma_2016_acta_comparative}. However, spaceborne images constantly suffer from low signal-to-noise ratio and high contrast; therefore, solving P$n$P becomes susceptible to severe feature outliers in addition to unknown feature correspondence and bad initial guesses. On the other hand, ML-based algorithms have recently emerged as a predominant approach to spaceborne monocular pose estimation \cite{sharma_2018_aero_pose, park_2019_aas, sharma_2020_taes_spn, pasqualettocassinis_scitech_2020_pose, chen_2019_iccvw_pose, proenca_2020_icra_urso, kaki_2023_jais_pose, park_2023_spnv2, garcia_2021_cvprw_lspnet}. They were in part encouraged by open-source datasets such as SPEED \cite{sharma_2020_taes_spn}, SPEED+ \cite{park_2022_speedplus} and many other comprising synthetic imagery of different spacecraft \cite{hu_2021_cvpr_wdr, musallam_2021_icipc_spark}. Several international competitions were also organized based on these datasets such as Satellite Pose Estimation Challenges (SPEC) \cite{kisantal_2020_taes_spec2019, park_2023_acta_spec2021} organized by Stanford's Space Rendezvous Laboratory (SLAB) and the European Space Agency.

\mysubsection{Bridging Sim2Real Gap} Addressing the sim2real gap has been a crucial topic in terrestrial robot vision \cite{peng_2018_cvprw_visda}. One approach is domain randomization \cite{tobin_2017_iros_domain_rand, zakharov_2019_iccv_deceptionnet} which aims to randomize various aspects of the synthetic images and rendering pipeline so that the target images appear as another randomized instances of the training set. Another is Unsupervised Domain Adaptation (UDA) \cite{bendavid_2007_nips_domain_adapt} which directly incorporates the unlabeled target domain images into the training phase. For example, in SPEC2021 \cite{park_2023_acta_spec2021} for spacecraft pose estimation across domain gap, the winners employed various adversarial training methods \cite{peng_2019_icml_dal, bousmalis_2016_nips_dsn} using unlabeled SPEED+ HIL domain images to close the sim2real gap. On a different line of work, Jawaid et al.~\cite{jawaid_2023_icra_eventsensing} employed event sensor images, which are agnostic to varying illumination conditions, to overcome the sim2real gap present in RGB and grayscale images.

The problem with many domain adaptation and adversarial training algorithms \cite{ganin_2016_jmlr_dann, sun_2016_eccv_deepcoral, tzeng_2017_cvpr_da} is that they require simultaneous access to both labeled training and unlabeled test data. This is not a realistic scenario for spaceborne applications considering the limited computational resources of satellite processors and the fact that flight images of the target spacecraft do not become available until RPO in space.

\mysubsection{Source-Free \& Test-Time Domain Adaptation} Another line of literature on domain adaptation that is conducive to operational constraints in space is source-free or test-time domain adaptation. Existing source-free algorithms leverage generative models for feature alignment \cite{li_2020_cvpr_modeladapt, kurmi_2021_wacv_domain_impress, yeh_2021_wacv_sofa} or pseudo-labeling and information maximization \cite{liang_2020_icml_hypo_transfer}. TENT \cite{wang_2021_iclr_tent} performs entropy minimization while updating only the affine parameters of the Batch Normalization (BN) layers \cite{ioffe_2015_icml_batchnorm}. The SPNv2 \cite{park_2023_spnv2} model for spacecraft pose estimation also utilized TENT on its satellite foreground segmentation task which has been shown to improve performance on pose estimation across domain gaps in SPEED+. However, methods based on entropy minimization require optimizations to be performed on batches of images in order to avoid trivial solutions, which could become computationally expensive on satellite avionics.

On the other hand, Test-Time Training (TTT) \cite{sun_2020_icml_ttt, liu_2021_nips_ttt++} trains an auxiliary Self-Supervised Learning (SSL) task from a shared feature encoder. During test time, the encoder is trained on SSL tasks, such as rotation prediction \cite{gidaris_2018_iclr_rot_pred} or image colorization \cite{larsson_2016_eccv_image_color}. However, TTT generally requires a hand-designed task or a large batch of negative sample pairs (e.g., contrastive learning \cite{chen_2020_icml_contrastive}).

Closely related to the presented methodology is Lu et al.~\cite{lu_2022_iros_slam_self_train} who self-train an object pose estimator NN using its predictions. In order to improve the accuracy of PLs and mitigate outliers, they take a SLAM-based approach and solve the Pose Graph Optimization (PGO) problem to enhance the consistency of PL across different images. However, PGO is first solved on \emph{all} images from the video stream, and the NN is fine-tuned on all pseudo-labeled images. The online training method presented in this work is instead much simpler and computationally efficient as it directly leverages the navigation filter already present on-board the satellite which can label the incoming flight images one at a time.

\section{Methodology} \label{sec:methodology}

\begin{figure}[!t]
\centering
    \centering
    \includegraphics[width=0.47\textwidth]{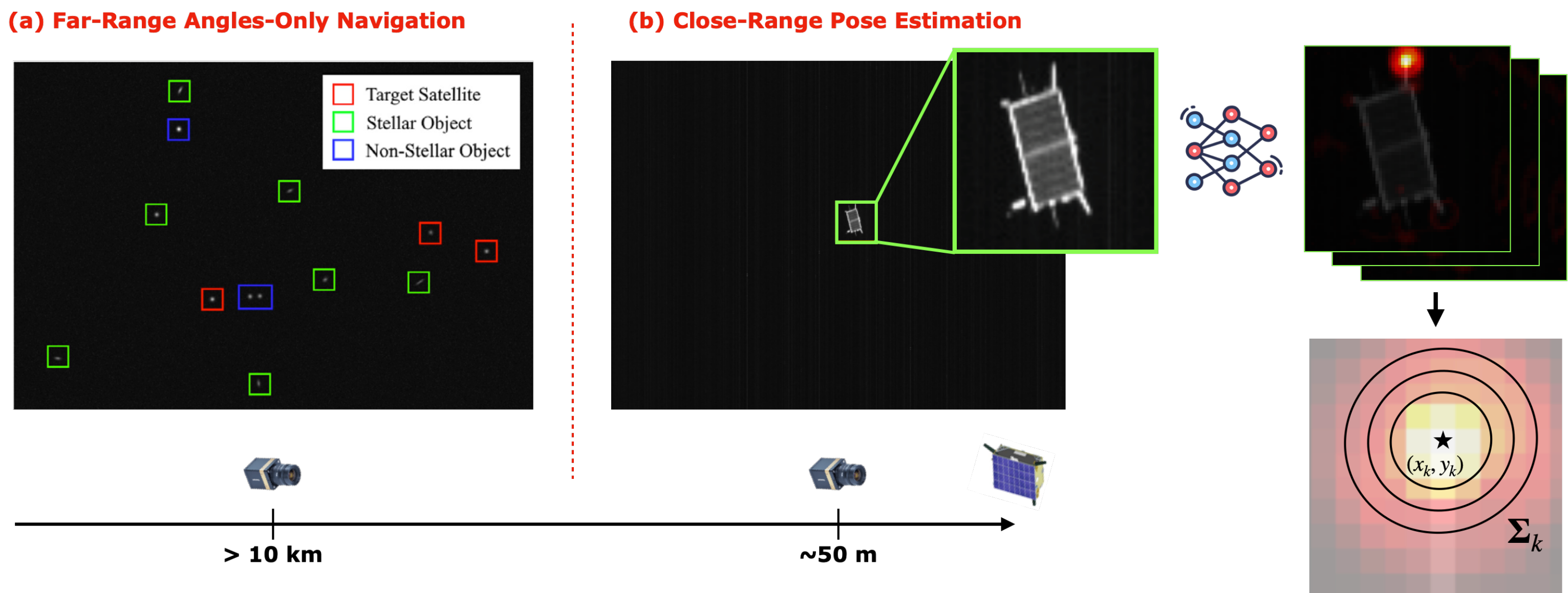}
    \caption{Visualization of (a) far-range AON \cite{kruger_2021_acta_multitrack} and (b) close-range pose estimation on a cropped image. The keypoint locations and uncertainties are extracted from heatmaps \cite{pasqualettocassinis_2021_acta_coupled}. Note that marked distances are specific to the presented imagery and may vary for camera FoV and target size.}
    \label{fig:operation}
\end{figure}

\mysubsection{Operational Scenario} In most vision-based RPO scenarios, the servicer spacecraft begins tracking the non-cooperative target at kilometers of separations. For example, Angles-Only Navigation (AON) \cite{damico_2013_jgcd_aon, sullivan_2021_jgcd_aon, kruger_2021_acta_multitrack} obtains bearing angle measurements of the target from a narrow Field-of-View (FoV) camera such as a star tracker which allows accurate tracking of the target's relative orbital state via nonlinear filtering. AON continues until the inter-spacecraft distance becomes small enough such that the target appears resolved in a camera, at which point the pose estimation algorithm kicks in (see Fig.~\ref{fig:operation}).

While the proper transition of orbital state estimates and uncertainties between far-range AON and close-range pose estimation remains an open problem, this work assumes a smooth handover between the two phases. Then, the relative position estimates from AON are used to crop the Region-of-Interest (RoI) around the target spacecraft. This is to ensure that pose estimation can be performed at a wide range of inter-spacecraft distances with minimum loss of information on the target's image features. This work also assumes that information about the target (e.g., CAD model, inertia moment) is available a priori in order to generate synthetic images and train NN (e.g., servicing missions). Finally, this work assumes that the servicer maintains an accurate estimate of its absolute orbital state and attitude through standard spacecraft avionics such as Global Navigation Satellite Systems (GNSS), star trackers and/or ground-based tracking and orbit determination.

\mysubsection{Pose Estimation Architecture} Given a cropped image $\mathcal{I} \in \mathbb{R}^{H \times W}$ of the RoI around the target spacecraft, this work considers a NN pose estimation architecture which outputs a set of $K$ heatmaps $\hat{\bm{\mathcal{H}}} = \{ \hat{\mathcal{H}}^{(k)} \}_{k=1}^K, \hat{\mathcal{H}}^{(k)} \in \mathbb{R}^{\frac{H}{4} \times \frac{W}{4}}$ centered at the 2D locations of known target surface keypoints. The keypoint locations $(x_k, y_k)$ are obtained from the heatmap peaks which are processed via Unbiased Data Processing (UDP) \cite{huang_2020_cvpr_udp}, and the uncertainties of keypoint measurements are approximated from the spread of heatmaps (see Fig.~\ref{fig:operation}). Specifically, the 2D covariance matrix $\bm{\Sigma_k}$ associated with the $k$-th keypoint is obtained via \cite{pasqualettocassinis_2021_acta_coupled}
\begin{align} \label{eqn:heatmap cov}
    \bm{\Sigma}_k = \begin{bmatrix} \text{c}_{xx} & \text{c}_{xy} \\ \text{c}_{yx} & \text{c}_{yy} \end{bmatrix}, ~ \text{c}_{xy} = \sum_{j = 1}^P w_j (x_j - x_k)(y_j - y_k)
\end{align}
where $w_j$ is the intensity of the $j$-th pixel, and $P$ is the number of pixels in the $k$-th heatmap. In this work, when evaluating the pose predictions from NN alone, EP$n$P \cite{lepetit_2008_epnp} is used to compute 6D pose solutions. Note that the NN predicts the heatmaps in a pre-defined order, so the feature 2D-3D correspondence is automatically established for any P$n$P algorithms.


\begin{figure}[!t]
\centering
    \centering
    \includegraphics[width=0.47\textwidth]{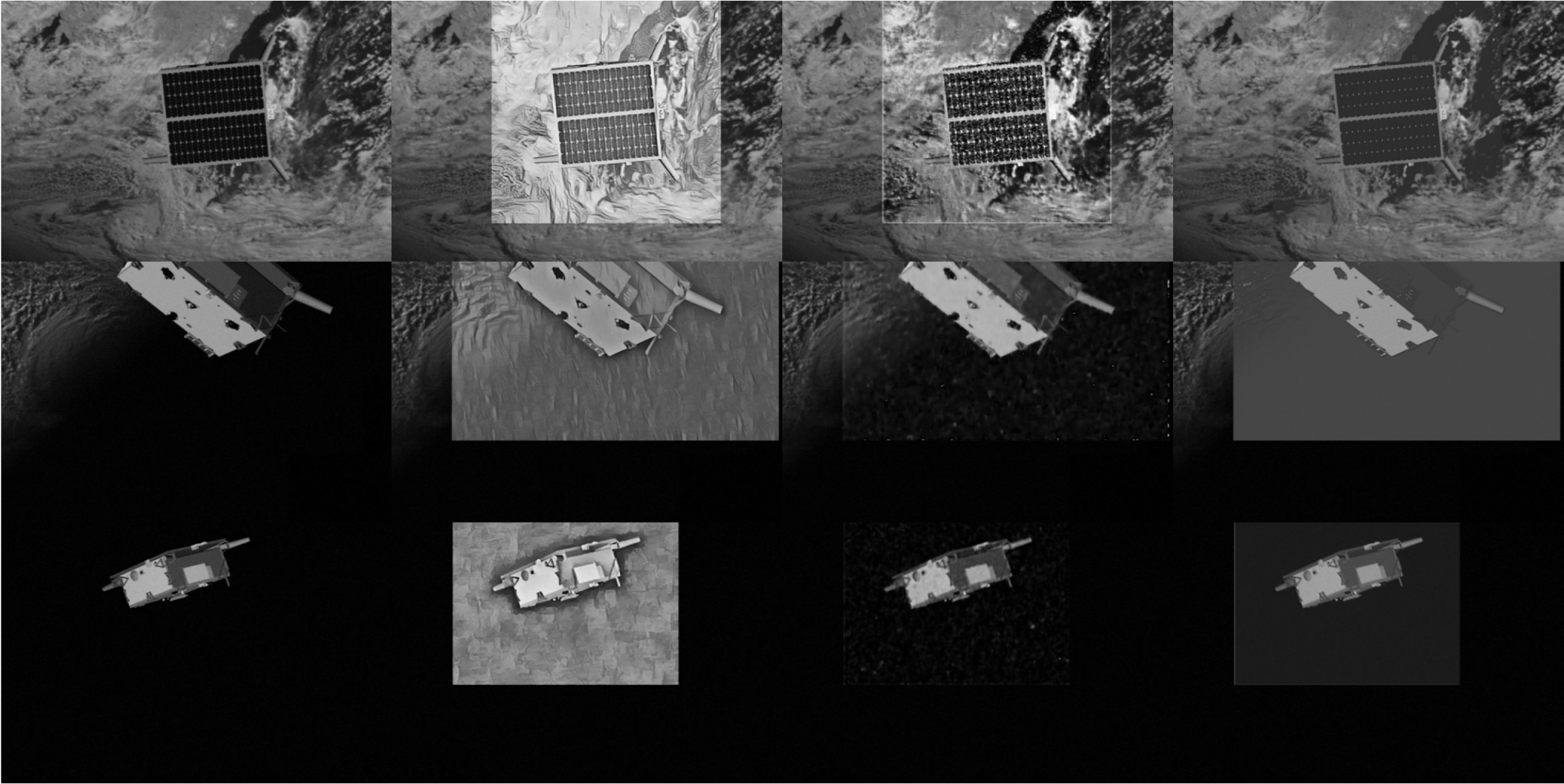}
    \caption{Image augmentations. From left to right: SPEED+ $\synthetic$ images; StyleAugment \cite{jackson_2019_cvpr_styleaug}; DeepAugment \cite{hendrycks_2021_iccv_imagenet_r}; RandConv \cite{xu_2021_iclr_randconv}.}
    \label{fig:augmentations}
\end{figure}

\mysubsection{Offline Robust Training} Recall that the goal is to perform OST on flight images using PLs generated from the navigation filter's state estimates (see Fig.~\ref{fig:pipeline}). Since the pose estimation NN is also a measurement module in the filter, it must be trained offline to be robust across domain gap using synthetic images. Note that HIL images are reserved for validation.

In order to train a robust pose estimation NN using only synthetic images, it is heavily augmented during the training phase. The goal of data augmentation is to amplify the NN inductive bias towards the spacecraft shape rather than local textures \cite{geirhos_2019_iclr_texture_bias} since the former is a feature common to all image domains. To this end, three additional versions of the synthetic images are created offline. The first is created with StyleAugment \cite{jackson_2019_cvpr_styleaug} which uses neural style transfer \cite{gatys_2016_cvpr_style_transfer} to randomly ``stylize'' the images. The second is created with DeepAugment \cite{hendrycks_2021_iccv_imagenet_r} which uses an image-to-image Convolutional NN (CNN) with randomly perturbed convolution weights. The third is created with RandConv \cite{xu_2021_iclr_randconv} which applies a convolution operation with random weights. See Fig.~\ref{fig:augmentations} which visualizes the aforementioned augmented versions of the SPEED+ $\synthetic$ image samples. Then, during training, one of four versions is randomly selected, which is further augmented by a random set of five augmentations from the Albumentations library \cite{buslaev_2020_albumentations} applied in the style of RandAugment \cite{cubuk_2020_nips_randaugment}.


\mysubsection{Adaptive Unscented Kalman Filter (AUKF)} This work uses an AUKF from Park \& D'Amico \cite{park_2023_jgcd_spnukf} for close-range navigation about a non-cooperative target. This section only reiterates key components on the filter.

The AUKF state vector is defined as $\bm{x} = [ \bm{\alpha}^\top ~~ \bm{p}^\top ~~ \bm{\omega}^\top ]^\top$. Here, $\bm{\alpha} \in \mathbb{R}^6$ denotes non-singular Relative Orbital Elements (ROE) \cite{koenig_2017_jgcd_stm} which tracks the orbital state of the target relative to the servicer; $\bm{p} \in \mathbb{R}^3$ is a Modified Rodrigues Parameters (MRP) vector tracking the step change in relative orientation of the target; and $\bm{\omega} \in \mathbb{R}^3$ tracks the relative angular velocity. The filter also keeps a reference quaternion vector $\bm{q}$ which tracks the relative orientation of the target. $\bm{q}$ is updated using $\bm{p}$ via the multiplicative approach of the Unscented Quaternion Estimator \cite{crassidis_2003_usque}. Note that ROE is used instead of a Cartesian vector as ROE is slowly varying in five of six elements and permits accurate state-transition matrices that can account for various perturbing forces in space such as Earth oblateness, drag \cite{koenig_2017_jgcd_stm} and solar radiation pressure \cite{guffanti_2019_jgcd_srp}.

The pose estimation NN provides keypoint measurements and their associated 2D covariance matrices approximated from the heatmaps via Eq.~\ref{eqn:heatmap cov}. The keypoint measurements form a measurement vector $\bm{y} = [ ~ x_1 ~ y_1 ~\cdots~ x_K ~ y_K ~ ]^\top \in \mathbb{R}^{2K}$ which is updated using ($\bm{r}$, $\bm{q}$) via projective transformation, where $\bm{r}$ is the relative position vector corresponding to the ROE state. The heatmap covariance matrices populate the 2 $\times$ 2 entries along the diagonals of the measurement noise covariance matrix $\bm{R}$. Any outlier is rejected based on the squared Mahalanobis distance of the AUKF innovation \cite{tweddle_2015_jgcd_mekf}.

The process noise covariance matrix $\bm{Q}$ is adaptively updated via Adaptive State Noise Compensation (ASNC) \cite{stacey_2021_taes_asnc}. ASNC uses analytical process noise models for both orbital and rotational motion \cite{stacey_2022_acta_covmodel, park_2023_jgcd_spnukf} based on the underlying continuous-time dynamics to update $\bm{Q}$ while ensuring that it remains positive semidefinite. It has been shown that ASNC significantly stabilizes the estimated filter states over time, especially the rotational motion when the heatmap measurements are noisy due to the domain gap.

\mysubsection{Online Supervised Training (OST)} In order to perform on-orbit OST, the heatmap PLs must be generated from the filter's most up-to-date state estimates. The process is visualized in Fig.~\ref{fig:pipeline}. First, at epoch $t$, the NN receives the flight image $\mathcal{I}_t$ and outputs $K$ heatmap predictions $\hat{\bm{\mathcal{H}}}_t$. During inference, the NN also computes and records the gradients. The heatmaps are converted to keypoint measurements and uncertainties (Eq.~\ref{eqn:heatmap cov}), and they are provided to AUKF which performs time update, measurement update, outlier rejection and process noise update via ASNC. The a posteriori state $\bm{x}_{t|t}$ is now in agreement with the underlying relative state for $\mathcal{I}_t$. Once all updates are complete, the state estimates are used to generate ``hard'' heatmap PLs $\bm{\mathcal{H}}_t^\text{pl}$ by drawing a Gaussian blob with $\sigma = 2$ [pix] around the keypoint locations. The standard deviation matches that used to generate ground-truth heatmaps during the offline training. In summary, 
\begin{align}
\mathcal{H}_t^{\text{pl}, (k)} = \texttt{GaussianBlob}(\Pi(\bm{x}_{t|t}, \bm{K}, \bm{k}_k), \sigma)
\end{align}
where $\Pi(\cdot)$ is a projective transformation function, $\bm{K}$ is the calibrated camera intrinsics, and $\bm{k}_k \in \mathbb{R}^3$ is the known 3D coordinate of the $k$-th keypoint.

Finally, once the heatmap PLs are created, the NN is trained with a Mean Squared Error (MSE) loss function, 
\begin{align}
  \mathcal{L}(\hat{\bm{\mathcal{H}}}_t, \bm{\mathcal{H}}_t^\text{pl}) = \frac{1}{K} \sum_{k=1}^K \| \hat{\mathcal{H}}_t^{(k)} - \mathcal{H}_t^{\text{pl},(k)} \|_F^2
\end{align}
The training only involves a single round of NN backpropagation on a single image to minimize the incurred computational effort. Note that by recording gradients from the earlier forward propagation, the NN optimization can be readily performed without additional passes. Moreover, even if $\hat{\bm{\mathcal{H}}}_t$ contains outliers, $\bm{\mathcal{H}}_t^\text{pl}$ are created from the filter state estimates for all keypoints regardless of outliers in raw predictions.

The key to preventing overfitting is to ensure that the servicer observes the target from a diverse set of directions between subsequent OST steps. This is equivalent to i.i.d.~sampling of training images from the spaceborne image domain, which is a core assumption for ML training. In reality, a satellite must follow prescribed approach trajectories with little fuel for a dedicated data collection phase. Instead, it can be placed in a passively safe relative trajectory around the target  \cite{montenbruck_2006_ast_eivectorsep} where it can observe the target from different directions. This means the images for OST must be selected occasionally to ensure there has been an enough change in viewing directions between subsequent OST steps. In this work, the training frequency is manually set to every 10 images; however, this can be triggered by other more sophisticated criteria such as sampling time and change of pose.

\mysubsection{Discussions} Unlike domain adaptation methods such as AdaBN \cite{li_2017_iclrw_adabn} and TENT \cite{wang_2021_iclr_tent} which only update the affine parameters of the BN layers, the proposed OST trains \emph{all} NN parameters via single rounds of backpropagation on single images. This is only feasible if the NN is small, computationally efficient and does not include BN layers. An interesting challenge is that as NN capacity decreases, it generalizes poorly to unseen test data \cite{simonyan_2015_vgg, szegedy2015_cvpr_inception, he2016_cvpr_resnet, tan_2019_icml_efficientnet}, adversarial attacks \cite{madry2018_iclr_towards}, common corruptions \cite{hendrycks2018_iclr_benchmarking} and out-of-distribution samples \cite{hendrycks_2021_iccv_imagenet_r}. While this work does not explore how to mitigate loss of robustness as NN gets smaller, the effect is simulated by prematurely ending the training after different numbers of training epoch and testing them individually for OST. 

Moreover, the proposed OST method requires that the filter converges before the online training begins. Prior work \cite{park_2023_jgcd_spnukf} showed through exhaustive experiments that the proposed AUKF is robust to errors in the servicer's absolute pose, initial $\bm{Q}$ and domain gap. In this work, the NNs with different epochs of training are also used to examine the filter's ability to converge, which is a core assumption of the proposed method.

\section{Experiments}

\mysubsection{Datasets} The offline robust training and OST are performed with the SPEED+ \cite{park_2022_speedplus} and SHIRT \cite{park_2023_jgcd_spnukf} datasets. The offline training is done with the SPEED+ $\synthetic$ imagery which consists of 60,000 images of the Tango spacecraft rendered with OpenGL. The robustness of NN alone is evaluated using 6,740 $\lightbox$ and 2,791 $\sunlamp$ labeled HIL images of the SPEED+. It is also evaluated on 25 flight images ($\prisma$) of the same target captured during the RPO of the PRISMA mission \cite{prisma_chapter, gill_2007_jgcd_prisma}. Recall \mbox{Fig.~\ref{fig:image domains}}.

\begin{figure}[!t]
\centering
\includegraphics[width=0.47\textwidth]{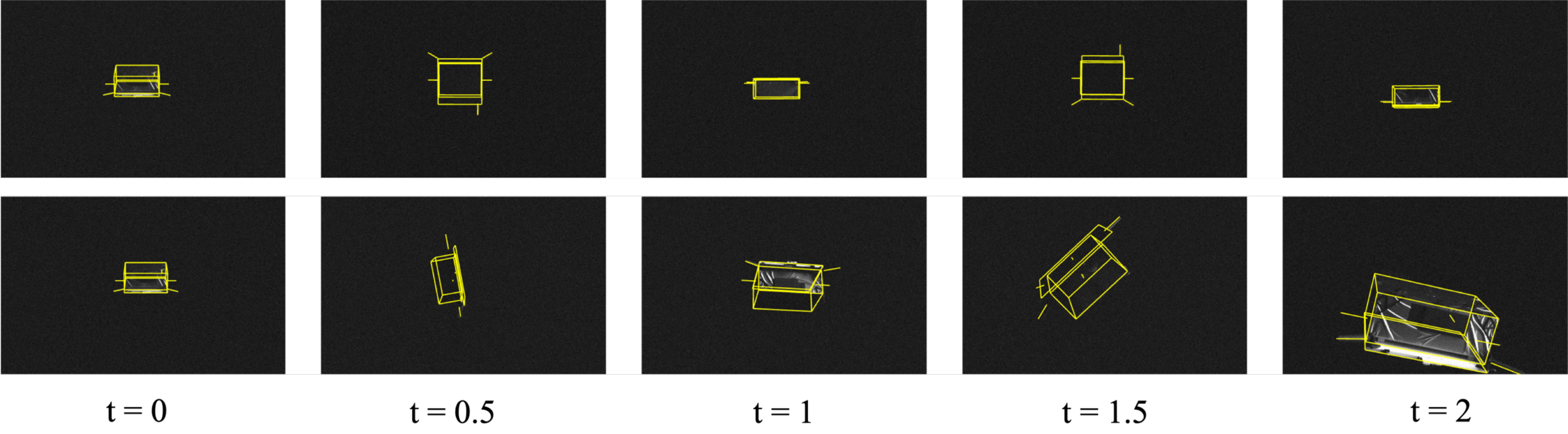}
\caption{Visualization of $\lightbox$ images from the SHIRT ROE1 (\emph{top}) and ROE2 (\emph{bottom}) trajectories with the target wireframe model projected with the ground-truth pose labels. Time ($t$) unit is in orbits.}
\label{fig:trajectories}
\end{figure}

When NN is integrated into AUKF, the filter's performance is evaluated on SHIRT which consists of $\synthetic$ and $\lightbox$ HIL images for two close-range rendezvous scenarios (ROE1 and ROE2) that last 2 orbits at 700 km altitude. With measurement intervals of 5 seconds, this amounts to 2,371 images per each trajectory. In ROE1, the servicer maintains a constant along-track separation while the target rotates about one axis. In ROE2, the servicer instead follows a spiral approach trajectory while the target rotates about two axes. Fig.~\ref{fig:trajectories} visualizes sample $\lightbox$ images from both trajectories. Naturally, ROE1 presents a more difficult scenario with limited variation in the target's view and distance.

\mysubsection{Training Details} This work uses ViTPose \cite{xu_2022_nips_vitpose} which leverages a Vision Transformer (ViT) \cite{dosovitskiy_2021_iclr_vit} backbone with 16 $\times$ 16 patches\footnote{The model pre-trained on ImageNet-1K \cite{deng_2009_cvpr_imagenet} is taken from HuggingFace \cite{wightman_2019_timm}: \texttt{timm/deit\_tiny\_patch16\_224.fb\_in1k}.}. ViT utilizes batch-agnostic Layer Normalization (LN) layers \cite{ba_2016_layernorm} which are conducive to the aforementioned operational constraint. The ViT backbone is followed by two upsampling convolution and LN layers which output $K$ = 11 heatmaps at 1/4 resolution. Considering limited computational resources, this work uses the smallest variant, ViTPose-T/16, which has 5.8M parameters. It is trained with the AdamW \cite{loshchilov_2019_iclr_adamw} optimizer, batch size 32, weight decay 0.1 and initial learning rate 0.001 which linearly warms up during the first epoch then decays with the cosine annealing rule \cite{loshchilov_2017_iclr_sgdr}. It is trained for 30 epochs on an NVIDIA RTX 4090 24GB GPU with PyTorch.

The training images are cropped around the target using the ground-truth bounding box and resized to 256 $\times$ 256. During training, the bounding boxes are enlarged and shifted by random factors to accommodate imperfect RoI detection from filter state estimates \cite{park_2019_aas}. During testing, the bounding box is enlarged by a constant factor before cropping. Recall Section \ref{sec:methodology} for details on data augmentation. Finally, motivated by recent flight heritages \cite{lockheed_gpu, aitech_gpu}, an NVIDIA Jetson Nano 4GB is used to measure the training/test time with the PyTorch C++ API. Equipped with a 128 CUDA core GPU, the Jetson Nano provides a restrictive GPU compute environment that is currently finding its application in LEO.

\mysubsection{Metrics} The accuracy of the predicted pose is reported as mean and standard deviation of translation ($E_\text{t}$) and orientation ($E_\text{q}$) errors, defined for the $i$-th sample as
\begin{align}
    E_\text{t}^{(i)} = \| \hat{\bm{t}}_i - \bm{t}_i \|_2, ~~ E_\text{q}^{(i)} = 2 \arccos |<\hat{\bm{q}}_i, \bm{q}_i>|
\end{align}
where $(\hat{\bm{t}}, \hat{\bm{q}})$ and $(\bm{t}, \bm{q})$ are predicted and ground-truth translation and quaternion vectors, respectively. In AUKF, the ROE state estimate $\hat{\bm{\alpha}}$ are first converted to $\hat{\bm{t}}$ \cite{damico_2010_phd}. Note that when evaluated on SPEED+ HIL domains, the calibrated metrics are used instead in order to incorporate the accuracy of labels recovered from the robotic facility \cite{park_2021_aas_tron}. Specifically, if $E_\text{t}^{(i)} / \| \bm{t} \| < \text{2.173 mm/m}$ or $E_\text{q}^{(i)} < 0.169^\circ$, the respective error is set to zero for that sample \cite{park_2023_acta_spec2021}. Finally, on SHIRT, the steady-state errors ($E_\text{t}^\text{ss}$, $E_\text{q}^\text{ss}$) of the filter state estimates are reported, which are averaged throughout the second orbit.

\section{Results}

\mysubsection{Offline Robust Training} First, the performance of offline robust training on ViTPose-T/16 is reported in Table \ref{table:backbone ablation}. Only $E_\text{q}$ are reported for brevity. For comparison, it also provides those of SPNv2 \cite{park_2023_spnv2}, a CNN which consists of the EfficientDet \cite{tan_2020_cvpr_efficientdet} backbone and a heatmap prediction head from the highest resolution feature level. Its training procedure is identical to that of ViTPose, different from what was reported in \cite{park_2023_spnv2}. All model backbones are pre-trained on ImageNet \cite{deng_2009_cvpr_imagenet}.

Table \ref{table:backbone ablation} reports that, unsurprisingly, SPNv2-B3 with 10.6M parameters performs the best on all image domains. However, the performance starts dropping as a smaller backbone is used. Replacing all BN layers with batch-agnostic Group Normalization (GN) layers \mbox{\cite{wu_2018_eccv_groupnorm}} also improves the performance at the expense of slightly slower inference. On the other hand, ViTPose with 6.2M parameters performs slightly worse than SPNv2-B0 on $\lightbox$ images with less than half of the memory and training/test time despite having more parameters. This is because ViT models are composed of more efficient linear layers. However, its performance on $\sunlamp$ and $\prisma$ are much worse than SPNv2. Unlike CNN, ViT operates on image patches with global position encoding which is known to lack inductive bias in modeling local visual structures compared to CNN models. Also considering the lack of regularization effect from BN, they contribute to degradation of robustness against particularly challenging image domains such as $\sunlamp$ which is characterized by harsh illumination conditions from direct lighting (see Fig.~\ref{fig:image domains}).

\begin{table}[t]
\caption{NN evaluated on SPEED+ and $\prisma$. The times are measured for single images and FP32 precision. Mean{~\footnotesize{(std.~dev.)}} of all samples over 5 training sessions with different random seeds are reported. The image domain names are abbreviated to the first three letters (e.g., \texttt{syn.} = \synthetic). See Fig.~\ref{fig:image domains}.}
\label{table:backbone ablation}
\centering
\tabcolsep=0.05cm
\begin{tabular}{lccccccccc}
\toprule
\multirow{2}{*}{Architecture} & \multirow{2}{*}{Norm} & \multirow{2}{*}{\makecell{Num. \\ Param.}} & \multirow{2}{*}{\makecell{Mem. \\ {[}MB{]}}} & \multicolumn{2}{c}{Time {[}ms{]}} & \multicolumn{4}{c}{$E_\text{q}$ [${}^\circ$]} \\
\cmidrule(lr){5-6} \cmidrule(lr){7-10}
& & & & Train & Test & \texttt{syn.} & \texttt{lig.} & \texttt{sun.} & \texttt{pri.} \\
\midrule 
SPNv2-B0 & BN & 4.1M & 269 & 491 & 57 & \repv{ 0.71 }{ 0.00 } & \repv{ 6.33 }{ 0.31 } & \repv{ 10.58 }{ 0.22 } & \repv{ 7.63 }{ 2.20 } \\ 
SPNv2-B0 & GN & 4.1M & 269 & 471 & 68 & \repv{ 0.73 }{ 0.02 } & \repv{ 5.28 }{ 0.13 } & \repv{ 9.13 }{ 0.31 } & \repv{ 2.92 }{ 2.06 } \\ 
SPNv2-B3 & BN & 10.6M & 453 & 817 & 75 & \repv{ 0.56 }{ 0.01 } & \repv{ 4.09 }{ 0.25 } & \repv{ 5.59 }{ 0.22 } & \repv{ 5.29 }{ 3.10 } \\ 
\midrule
ViTPose-T/16 & LN & 6.2M & 138 & 242 & 24 & \repv{ 1.01 }{ 0.02 } & \repv{ 8.00 }{ 0.15 } & \repv{ 16.89 }{ 0.28 } & \repv{ 13.11 }{ 1.99 } \\ 
\bottomrule
\end{tabular}
\end{table}

\mysubsection{Online Supervised Training} As discussed in Section \ref{sec:methodology}, decreasing the NN capacity leads to worse performance on unseen test data. In order to simulate loss of accuracy due to different NN sizes, ViTPose-T/16 training is ended prematurely at different epochs. These models are integrated into the filter (recall Fig.~\ref{fig:pipeline}) and simulated on the SHIRT $\lightbox$ ROE1 and ROE2 trajectories that have not been seen by the network during offline training. The network is trained with the same optimizer (AdamW) and weight decay but with a constant learning rate of $1 \times 10^{-5}$. For this experiment, the ViTPose models trained in PyTorch are serialized, and OST is implemented using PyTorch C++ API. The AUKF is also implemented in C++.

\begin{figure}[!t]
\centering
\includegraphics[width=0.48\textwidth]{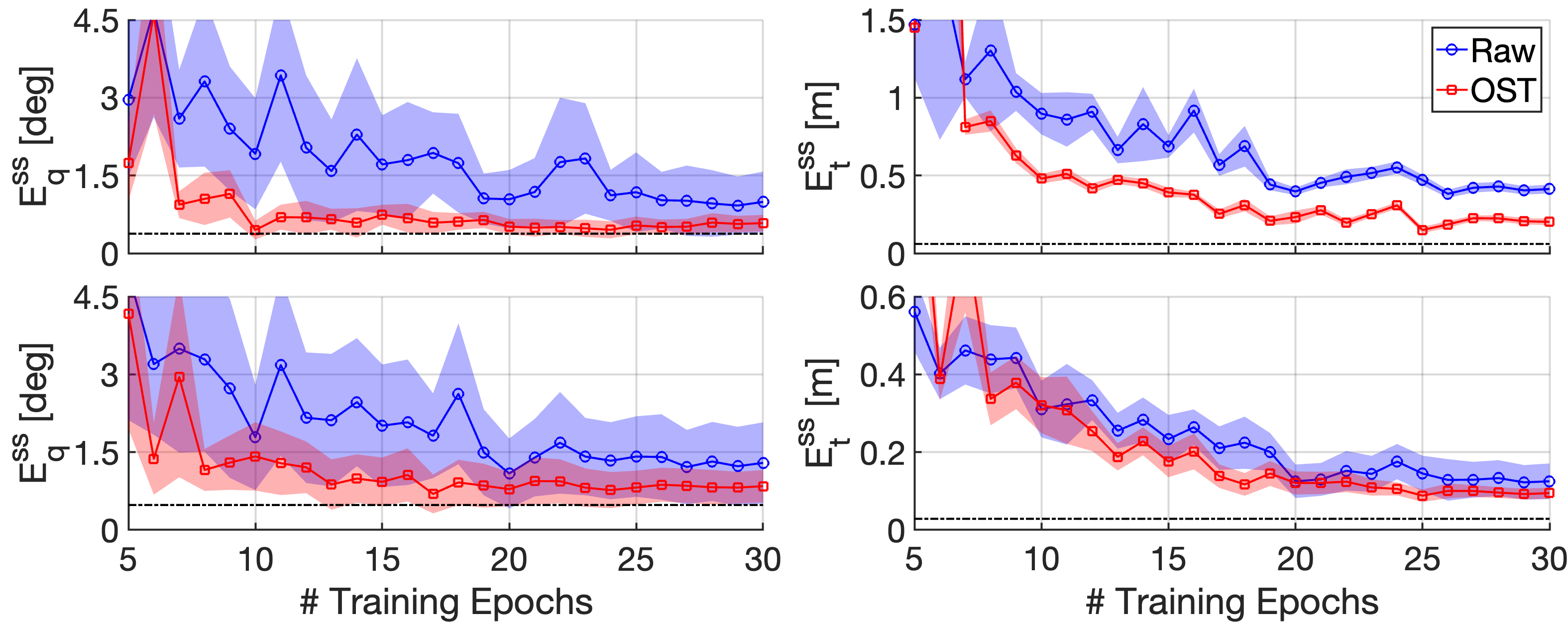}
\caption{AUKF steady-state errors without OST (\emph{raw}) and with OST on SHIRT ROE1 (\emph{top}) and ROE2 (\emph{bottom}) trajectories when ViTPose-T/16 is trained for different number of epochs. The shaded regions denote standard deviation. Black horizontal lines indicate baseline steady-state errors of AUKF with fully-trained (30 epochs) model evaluated on corresponding SHIRT $\synthetic$ trajectories.}
\label{fig:ukf_nn_perf}
\end{figure}

First, Fig.~\ref{fig:ukf_nn_perf} reports the mean and standard deviation of $E_\text{q}^\text{ss}$ and $E_\text{t}^\text{ss}$. ViTPose was trained for as few as 5 epochs in the 30 epochs training schedule. It can be seen that OST brings down the steady-state errors for both ROE1 and ROE2 compared to those without OST, and the improvement is consistent for the models trained longer than 8 epochs for all metrics. Note that the improvement is more dramatic for ROE1, where $E_\text{q}^\text{ss}$ can be seen to match the baseline steady-state error on $\synthetic$ trajectories at just 10 epochs of offline training. Moreover, the smaller standard deviations also indicate that the filter state estimates are more stable despite increased noise of NN prediction due to sub-optimal training.

\begin{figure}[!t]
\centering
\begin{subfigure}[b]{0.24\textwidth}
    \centering
    \includegraphics[width=\textwidth]{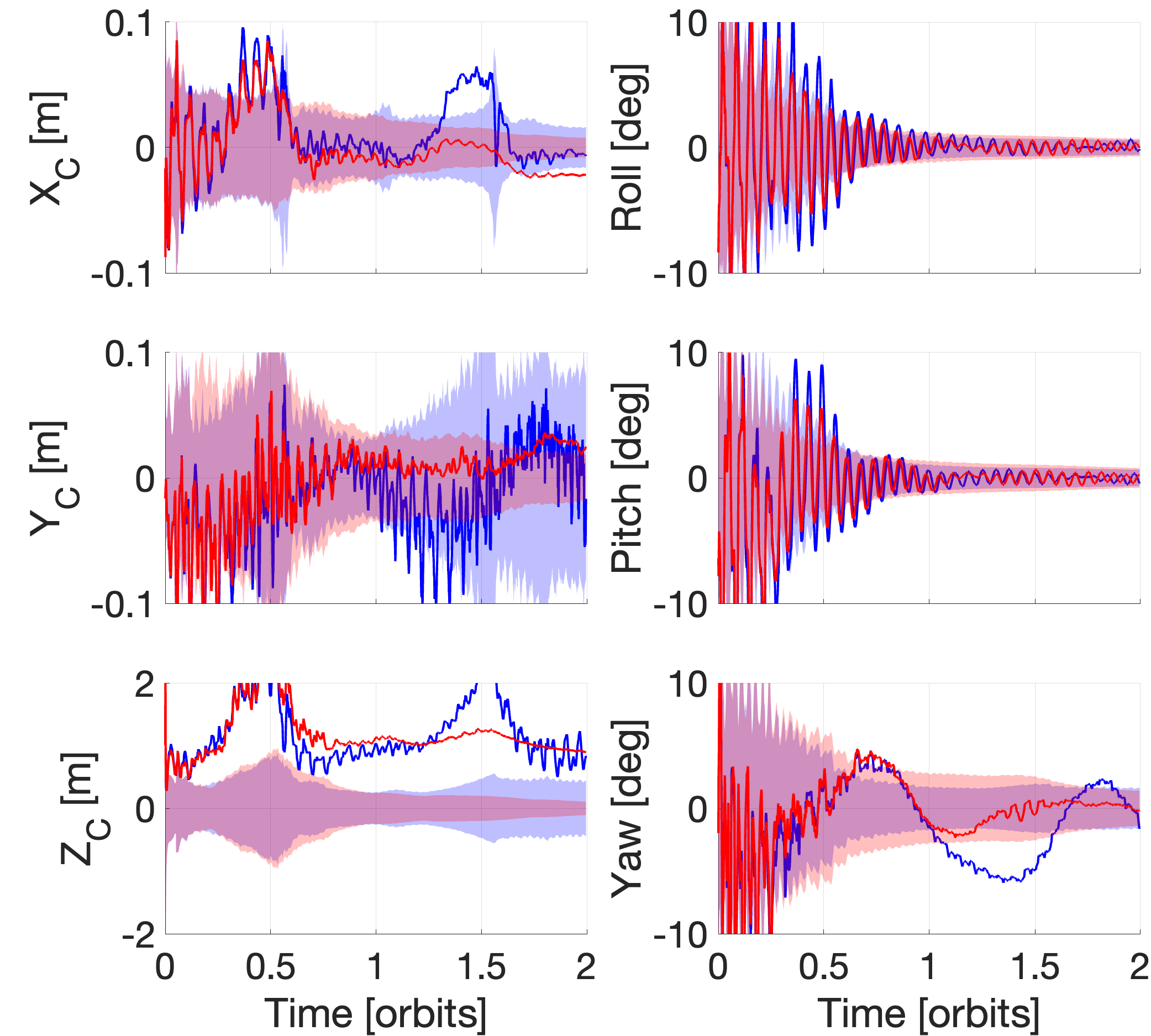}
\end{subfigure}
\begin{subfigure}[b]{0.24\textwidth}
    \centering
    \includegraphics[width=\textwidth]{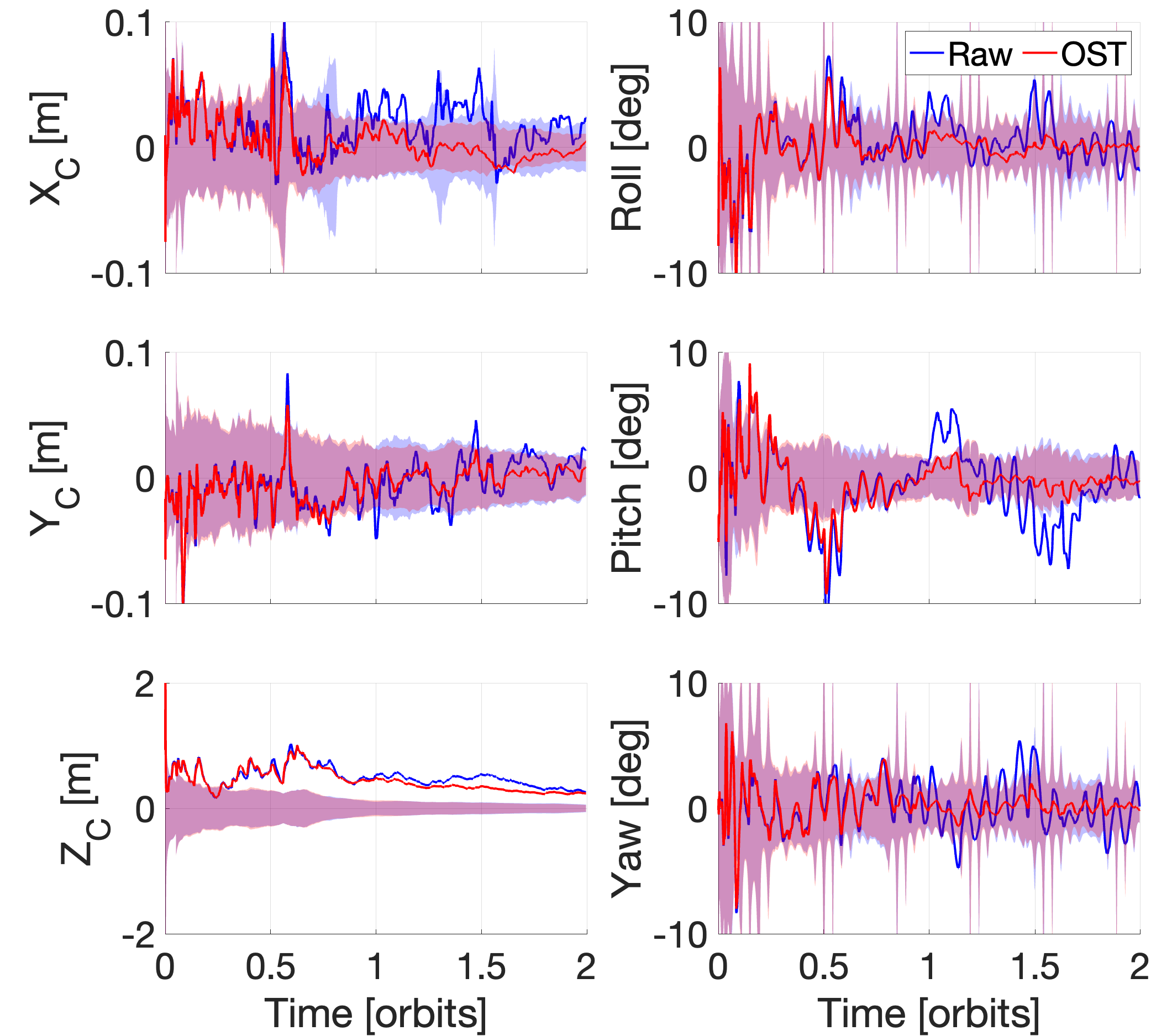}
\end{subfigure}
\caption{AUKF convergence without OST (\emph{raw}) and with OST on SHIRT ROE1 (\emph{left}) and ROE2 (\emph{right}) trajectories when ViTPose-T/16 is trained sub-optimally for 8 epochs. The shaded regions denote 3-$\sigma$ uncertainties.}
\label{fig:ukf convergence}
\end{figure}

Based on the minimum number of epochs necessary for improvement after OST, Fig.~\ref{fig:ukf convergence} shows the AUKF convergence when NN is only trained offline for 8 epochs. The filter state vector is converted to relative pose vectors expressed in the camera frame, and the associated covariance matrices are computed from the state uncertainties via unscented transform. The improvement of AUKF performance is visible especially during the second orbit and for ROE1. Note that $E_\text{t}^\text{ss}$ is dominated by the translation along the $z$-axis which coincides with the camera boresight. Since it is the least observable element, especially for ROE1 where the target remains at roughly constant distance (recall Fig.~\ref{fig:trajectories}), the error along $z$-axis still remains significant even with OST over just 2 orbits.

\begin{figure}[!t]
\centering
\begin{subfigure}[b]{0.24\textwidth}
    \centering
    \includegraphics[width=\textwidth]{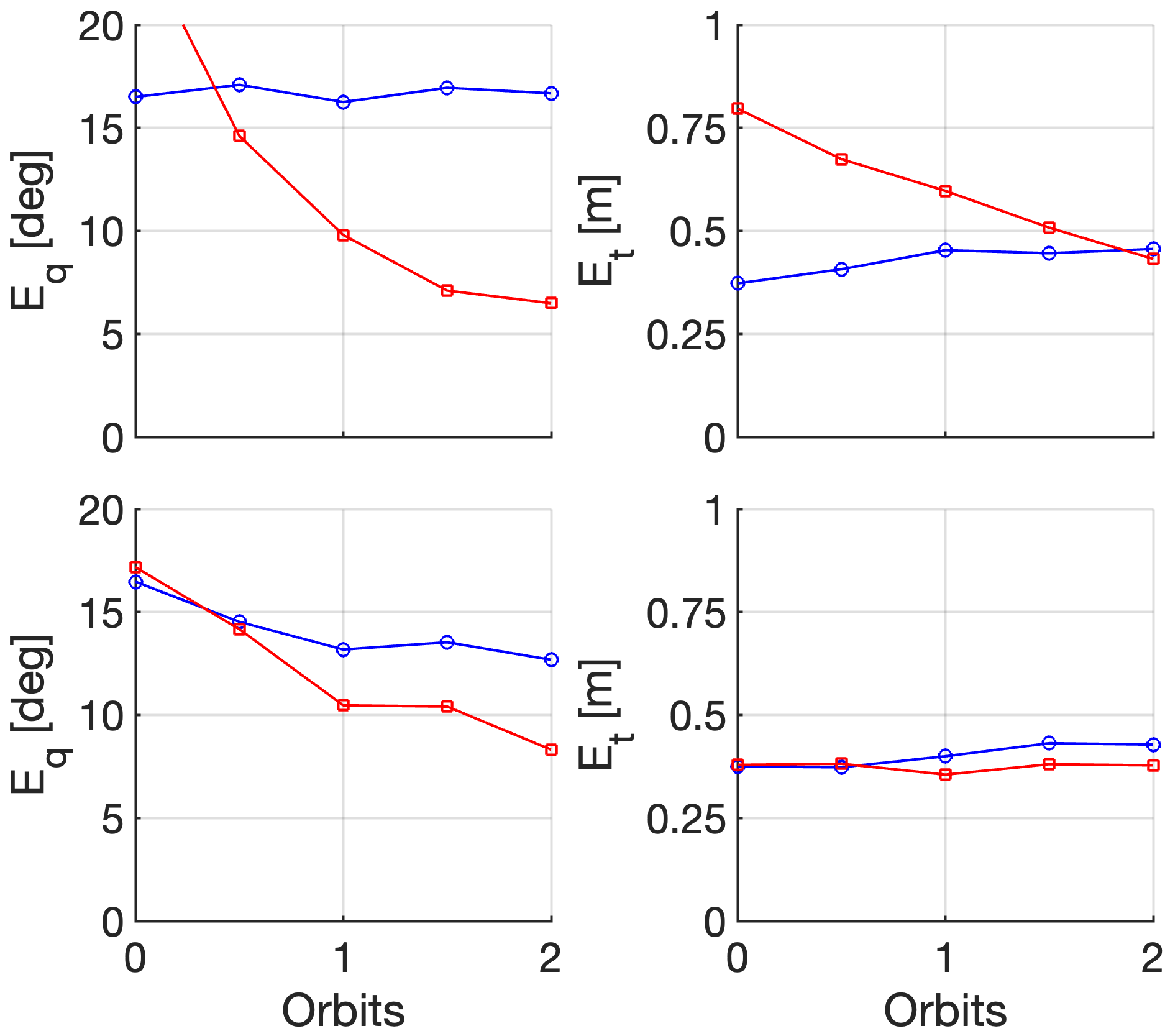}
    \caption{$E$ = 10}
\end{subfigure}
\begin{subfigure}[b]{0.24\textwidth}
    \centering
    \includegraphics[width=\textwidth]{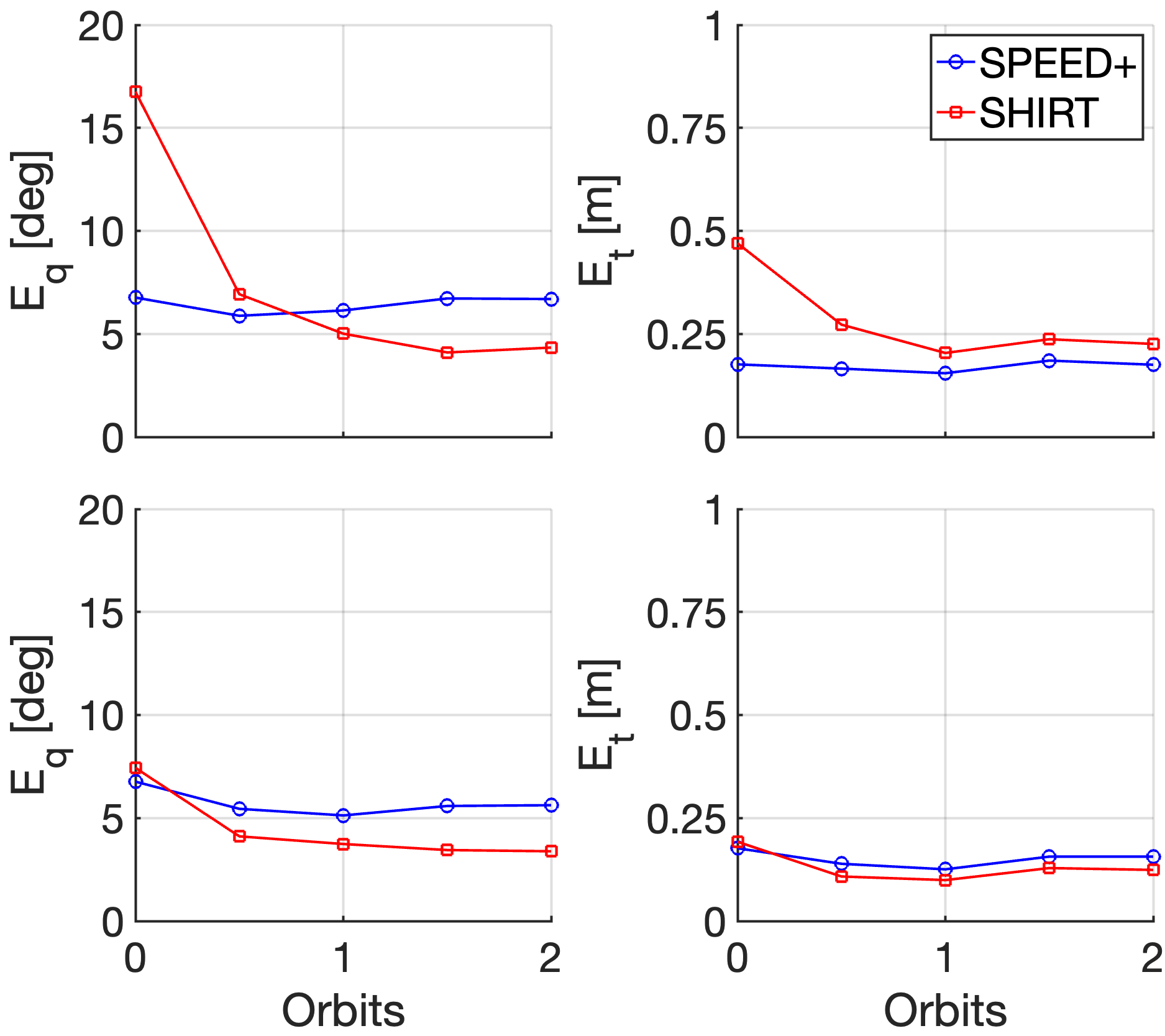}
    \caption{$E$ = 30}
\end{subfigure}
\caption{Performances of ViTPose-T/16 alone sub-optimally trained offline for $E$ epochs when evaluated on HIL $\lightbox$ images of SPEED+ and SHIRT after OST on ROE1 (\emph{top}) and ROE2 (\emph{bottom}).}
\label{fig:nn_perf_no_ukf}
\end{figure}

Finally, Fig.~\ref{fig:nn_perf_no_ukf} investigates the generalizability of ViTPose trained offline for 10 and 30 epochs. Each NN is trained online via OST on SHIRT, and it is evaluated in isolation on both SHIRT and SPEED+ HIL $\lightbox$ images, where the latter assesses its \emph{general} performance on the image domain as a whole. First, Fig.~\ref{fig:nn_perf_no_ukf} indicates that performing OST on both trajectories monotonically brings down $E_\text{q}$ and $E_\text{t}$ of ViTPose on SHIRT images. However, the errors remain about the same level on SPEED+, which suggests that the model is indeed overfitting to the SHIRT images but not to the point where it hurts the model's generalization capability on all $\lightbox$ domain images. In fact, $E_\text{q}$ on SPEED+ \emph{decreases} as OST progresses on ROE2. Given that ROE2 images are captured from diverse directions (recall Fig.~\ref{fig:trajectories}), it suggests that geometric diversity in OST training images is not only crucial to prevent ``catastrophic forgetting'' but also improve its pose estimation at all views.

\mysubsection{Limitations} The main limitation of the OST experiments is that the SHIRT trajectories are limited to 2 orbits in LEO and therefore do not allow investigating the long-term behavior of the coupling between OST and AUKF. For example, Fig.~\ref{fig:ukf convergence} indicates that there is a bias in translation along the $z$-axis. It is left to future work to understand how eventual biases or instabilities of AUKF affect OST on the long term.



\section{Conclusion}

This work presented an Online Supervised Training (OST) procedure to close the domain gap between synthetic training images and spaceborne flight images of a spacecraft pose estimation Neural Network (NN). The NN is integrated as a measurement module into the satellite's Adaptive Unscented Kalman Filter (AUKF), and the filter's estimates of the target's relative orbital and attitude motion are used to create Pseudo-Labels (PL) for OST. In order to comply with limited computational resources on-board the spacecraft, OST is performed with a single backpropagation on each flight image. The NN is trained offline on synthetic images to be robust on unseen flight images and ensure the quality of PL. The experiments on representative rendezvous trajectories with Hardware-In-the-Loop (HIL) images show that OST improves not only the AUKF steady-state errors but also the NN performance on the HIL domain as a whole provided that it is trained on diverse views of the target. The statement also applies to smaller NNs with weaker performance which was verified by prematurely ending the NN offline training and testing it for OST.

In the future, the proposed OST will be evaluated on much longer trajectories with HIL images in order to properly investigate the long-term coupling effect of OST and AUKF. Furthermore, OST will be improved by better reflecting the AUKF state uncertainties and incorporating more rigorous criteria to schedule OST during proximity operations.

\bibliographystyle{IEEEtran}
\bibliography{IEEEabrv, reference}

\end{document}